# Hybrid Machine Learning Model for Forest Height Estimation from TanDEM-X and Landsat Data

Islam Mansour, *Member, IEEE*, Ronny Hänsch, *Senior Member, IEEE*, Irena Hajnsek, *Fellow, IEEE*, and Konstantinos Papathanassiou, *Fellow, IEEE*

***Abstract*—Integrating machine learning (ML) with physical models (PM) has emerged as a promising way of retrieving geophysical parameters from remote sensing data. In this context, a ML model for estimating forest height from TanDEM-X interferometric coherence measurements has recently been proposed, that constrains the learning process through a PM. While the features used for training and inversion where selected to ensure the physical consistency of the solutions, they could not resolve all height / structure and baseline / terrain slope ambiguities in the data. To improve this, the extension of the feature space with optical Landsat data is proposed able to provide complementary information on forest type or structure. The extended model is applied and validated on several TanDEM-X acquisitions over the Gabonese Lopé national park site and assessed against airborne LiDAR measurements. Results show a 13.5% reduction in RMSE and a 16.6% reduction in MAE compared to the original hybrid model, confirming the added value of multispectral inputs.**

## I. INTRODUCTION

POLARIMETRIC SAR Interferometry, i.e., interferometric measurements with polarimetric and spatial baseline diversity, is an established remote sensing technique for obtaining continuous forest height estimates of significant accuracy at large spatial scales [1], [2], [3]. The inversion approaches are first-line model-based, exploring the inherent sensitivity of interferometric measurements to the 3D distribution of scatterers within forests. The achieved performance critically depends on the parameterization of the inversion problem. Accurate and generic parameterization requires an adequate observation space to achieve a balanced inversion problem. Simplified or constrained model parameterizations without adequate observation space result in compromised inversion performances.

Received 19 February 2026; accepted 7 May 2026. This work was supported by the DeepSAR Research Project, funded by Helmholtz AI under the Helmholtz Association of German Research Centers (HGF). (*Corresponding author: Islam Mansour.*)

Islam Mansour and Irena Hajnsek are with the Microwaves and Radar Institute, German Aerospace Center (DLR), 82234 Weßling, Germany, and also with the Chair of Earth Observation and Remote Sensing, Institute of Environmental Engineering, ETH Zürich, 8093 Zürich, Switzerland (e-mail: islam.mansour@dlr.de; irena.hajnsek@dlr.de).

Konstantinos Papathanassiou and Ronny Hänsch are with the Microwaves and Radar Institute, German Aerospace Center (DLR), 82234 Weßling, Germany (e-mail: kostas.papathanassiou@dlr.de; ronny.haensch@dlr.de).

Machine learning (ML) inversion approaches have been developed in the last few years as an alternative to exploring the potential of ML to identify patterns and correlations in complex, multi-dimensional space, even when explicit (physical) relations are not available. ML allows the combining of different data types into a single framework, including SAR-derived interferometric coherence and optical remote sensing features for forest height estimation [4], [5], [6]. However, despite their unquestioned potential, current ML models for forest height estimation often suffer from limited interpretability and generalization.

One way to overcome these disadvantages has been proposed in the context of hybrid modeling approaches that integrate ML and physical/mathematical models [7], [8]. This allows to combine the expressiveness of data-driven methods with the interpretability and generalization of physical models. In this context, a recently published study established a hybrid inversion framework combining ML and physical modeling components to estimate forest height from TanDEM-X interferometric coherence measurements [8].

Accordingly, the vertical reflectivity profile is represented by a Legendre series expansion, with its coefficients predicted as a function of input features via a multilayer perceptron (MLP). The input features—interferometric volume coherence, terrain-corrected vertical wavenumber, incidence angle, and terrain slope—were selected to allow an optimum parameterization of the PM. The resulting reflectivity profile was then used within the physical model to estimate forest height.

The advantages of this approach are twofold. First, it enables forest height estimation—even without the full observation space required for a traditional inversion—by allowing the MLP to infer the reflectivity profile from limited input features. Second, it reduces the number of required acquisitions by leveraging physical constraints to generalize across vertical wavenumbers, even when trained on a limited subset.

However, while the selected features used for training and inversion allow an optimum parameterization of the physical model and so ensure the physical consistency of the solutions, they appear are not enough to resolve all height / structure and baseline / terrain slope ambiguities. This is because of the insufficiency of the training data to sample all ambiguous conditions especially when it comes to changes in the underlying forest structure or terrain slopes where the modulation of the interferometric coherence is primarily induced by the change in interferometric coherence (e.g. the vertical wavenumber). As a result, while physically consistent, the earlier method lacked adaptability to different forest and terrain conditions, which contributed to the relatively high RMSE values reported.

This study addresses that limitation by extending the feature space with multispectral Landsat data, which provide complementary information on vegetation variability not captured by InSAR observables or geometry-related parameters. In this way, the ML component is exploited to learn additional mappings for the Legendre coefficients beyond those constrained by the PM, making the framework more adaptable to forest structure diversity and enhancing the solution space available for inversion.

## II. Methodology

The measured interferometric coherence $\tilde{\gamma}_{\mathrm{Vol}}(\kappa_z)$ can be expressed as [18], [21]:

$$\tilde{\gamma}_{\mathrm{Obs}}(\kappa_z) = \tilde{\gamma}_{\mathrm{Tmp}}\,\tilde{\gamma}_{\mathrm{Rg}}\,\tilde{\gamma}_{\mathrm{Sys}}\,\tilde{\gamma}_{\mathrm{Vol}}(\kappa_z) \qquad (1)$$

where $\tilde{\gamma}_{\mathrm{Vol}}(\kappa_z)$ accounts for volume decorrelation related to the forest canopy, while $\tilde{\gamma}_{\mathrm{Tmp}}, \tilde{\gamma}_{\mathrm{Rg}}, \tilde{\gamma}_{\mathrm{Sys}}$ represent temporal, range, and system decorrelations, respectively [13]. The volume decorrelation is defined as [14], [15]:

$$\tilde{\gamma}_{\mathrm{Vol}}(\kappa_z) = e^{i\kappa_z z_0} \frac{\int_0^{h_v} f(z)\, e^{i\kappa_z z} dz}{\int_0^{h_v} f(z) dz} \qquad (2)$$

where the vertical reflectivity profile $f(z)$ represents the vertical distribution of scattering elements and $h_v$ the forest height. The vertical wavenumber $\kappa_z$ (in rad/m), expresses the sensitivity of the interferometric phase to height changes [16]:

$$\kappa_z = m \frac{2\pi}{\lambda} \frac{\Delta\theta}{\sin(\theta_0 + \alpha)} \qquad (3)$$

where $\theta_0$ is the radar look angle, $\Delta\theta$ the look angle difference induced by the spatial baseline, $\lambda$ the wavelength, and $\alpha$ the ground range terrain slope. The factor m depends on the acquisition mode (m = 2 for monostatic, m = 1 for bistatic). The sensitivity of phase-to-height changes can also be expressed by the height of ambiguity (HoA = $2\pi/\kappa_z$). When $f(z)$ is known, Eq. (2) is reduced to a balanced and unique inversion problem, allowing the forest height estimation.

In order to obtain an estimate for the vertical reflectivity profile, $f(z)$ is first expressed in terms of a polynomial series expansion using the Legendre polynomials $P_n(z)$ as [17], [18]:

$$f(z, a_n) = \sum_{n=1}^{N} a_n P_n(z) \qquad (4)$$

where $a_n$ are the associated coefficients, and $N$ is the number of terms in the expansion truncated to a relatively small number. After this, an ML algorithm is used to predict the underlying vertical reflectivity profile $f(z, a_n)$ expressed in terms of the Legendre series expansion (see Eq. 4) as a function of a set of input features. A purely data-driven MLP directly regressing forest height from TanDEM-X InSAR features was previously investigated in [7]. In contrast, the proposed framework constrains the learning process through the physically-based formulation in Eq. (2), improving physical consistency and generalization. The predicted vertical reflectivity profile $f(z, a_n)$ is then used in Eq. (2) to estimate forest height. Figure 1 shows a conceptual representation of the architecture and functionality of the proposed hybrid model in training and inference.

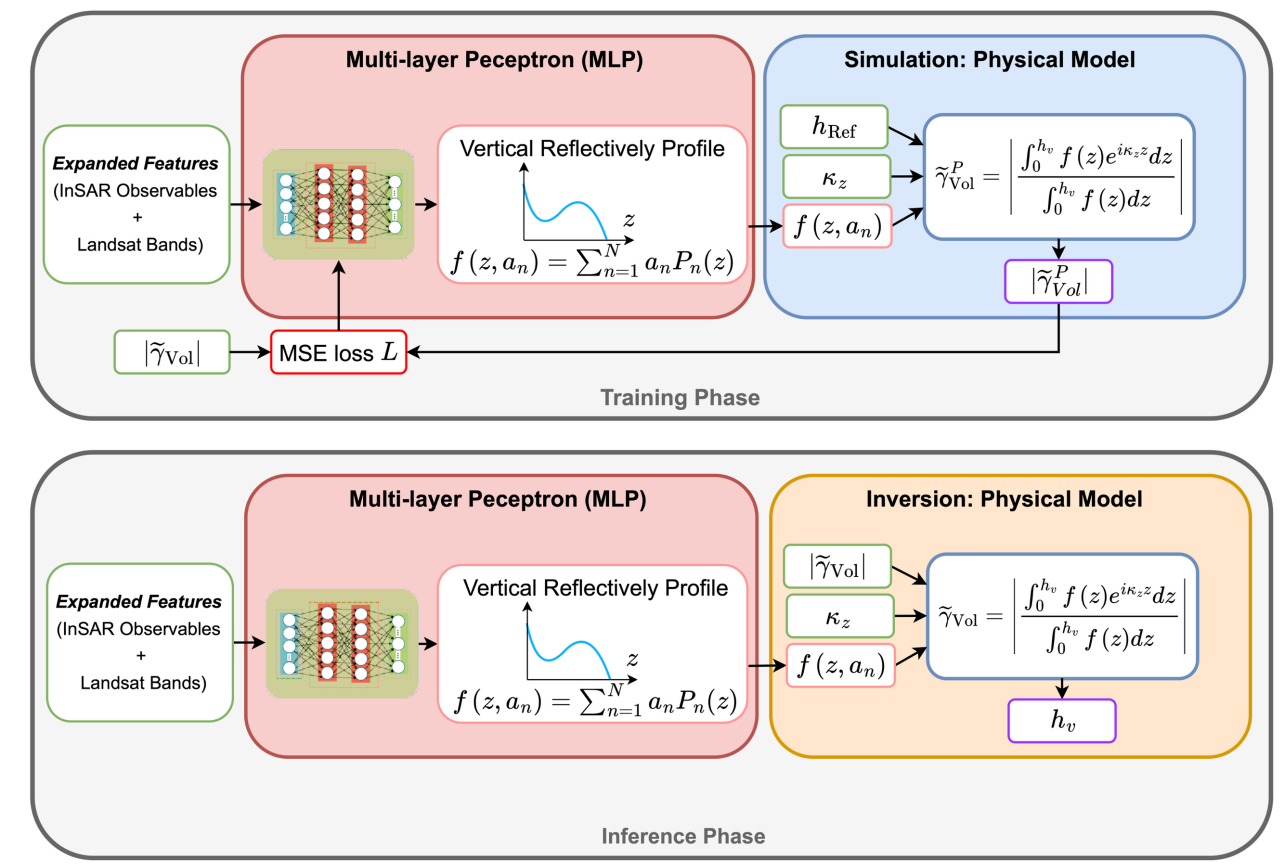

**Fig. 1.** Conceptual architecture and functionality of the hybrid model in the training (top) and inference phase (bottom).

**Table I:** Models and associated number of Legendre coef. and features used for training (as shown in Fig 1, top).

| Model | Coef. | ML Input Features | PM Inputs |
|---|---|---|---|
| C | 7 | $\kappa_z$, $\lvert\tilde{\gamma}_{\mathrm{Vol}}(\kappa_z)\rvert$, $\theta_0, \theta_{\mathrm{loc}}$, $\alpha$ | $\kappa_z, h_{Ref}, f(z, a_n)$ |
| D | 7 | $\kappa_z$, $\lvert\tilde{\gamma}_{\mathrm{Vol}}(\kappa_z)\rvert$, $\theta_0, \theta_{\mathrm{loc}}$, $\alpha$, $Red, NIR, SWIR1, SWIR2$ | $\kappa_z, h_{Ref}, f(z, a_n)$ |

The model with the best performance in [8] used seven Legendre polynomials (N=7) to define the vertical reflectivity profiles and three TanDEM-X acquisitions (two ascending and one descending) for training. The input features included the interferometric volume coherence $\tilde{\gamma}_{Vol}(\kappa_z)$, the terrain corrected vertical wavenumber $\kappa_z$, the incidence angle $\theta_0$, and the terrain slope in the range direction $\alpha$. This configuration is referred to as Model C. To improve its performance, the input feature space was extended by incorporating four Landsat spectral bands: Band 4 (Red, 0.630–0.680 µm), Band 5 (Near-Infrared [NIR], 0.845–0.885 µm), and Bands 6 and 7 corresponding to Short-Wave Infrared 1 (SWIR1, 1.560–1.660 µm) and Short-Wave Infrared 2 (SWIR2, 2.100–2.300 µm), respectively. This extended configuration is referred to as Model D.

To ensure consistency and comparability with Model C, the same basic setup was retained for Model D, including preprocessing steps, dataset split ratios, and optimization settings, as described in [8]. Moreover, the same training strategy used for Model C was applied to Model D. The models, their input features, and configurations are summarized in Table I. The only difference lies in including the four Landsat spectral bands in the input feature space.

**Table II**: TanDEM-X Data Sets. Site, Scene ID's Height of Ambiguity [m] (HoA), Nominal incidence angle [°]. The HoA sign indicates the orbit direction: positive for ascending, negative for descending orbits.

| No | Site | Scene ID | HoA [m] | $\theta$ [°] |
|---|---|---|---|---|
| 1 | Lopé | TDM1_SAR__COS_BIST_SM_S_SRA_20190610T173107_20190610T173115 | 52.45 | 46.18 |
| 2 | Lopé | TDM1_SAR__COS_BIST_SM_S_SRA_20160125T173041_20160125T173048 | -65.22 | 44.44 |
| 3 | Lopé | TDM1_SAR__COS_BIST_SM_S_SRA_20111002T045625_20111002T045633 | 86.34 | 46.08 |
| 4 | Lopé | TDM1_SAR__COS_BIST_SM_S_SRA_20121226T045626_20121226T045634 | 94.89 | 45.10 |
| 5 | Lopé | TDM1_SAR__COS_BIST_SM_S_SRA_20121215T045627_20121215T045635 | 95.41 | 46.68 |

## III. Study Area And Dataset

This study focuses on the Lopé site within the homonymous National Park in Gabon. The area comprises savannah and subtropical forest stands with diverse species composition and density. Tree heights exceed 50 m in many locations. The terrain is hilly, with local slopes sometimes even steeper than 20°.

The set of TanDEM-X images used is summarized in Table II. From the TanDEM-X CoSSC products, and following standard pre-processing steps, the volume decorrelation $\tilde{\gamma}_{\mathrm{Vol}}(\kappa_z)$, the terrain corrected vertical wavenumber $\kappa_z$, and the local incidence angle $\theta_0$ are derived. Terrain correction was performed using the 30 m Copernicus DEM [19] .

Four Landsat spectral bands (Red, NIR, SWIR1, and SWIR2) were derived from the circa 2019 cloud-free composite of the Hansen Global Forest Change v1.7 product [20], primarily based on Landsat-8 imagery, with fallback to the nearest cloud-free year (2010–2015).

Reference forest height data were derived from full-waveform LiDAR collected by NASA's LVIS sensor during the AfriSAR 2016 campaign [21]. The RH98 metric, representing the 98th percentile height, was used as the reference forest height ($h_{\mathrm{Ref}}$= RH98). All datasets were resampled and georeferenced to a standard spatial resolution of 20 × 20 m$^2$.

## IV. Implementation And Results

Both models were trained using the same three TanDEM-X acquisitions — two ascending and one descending — corresponding to scenes 1, 2, and 5 in Table III. These acquisitions were chosen because they span a range of Height of Ambiguity (HoA) values (52.45 m, −65.22 m, and 95.41 m), which ensures different sensitivities of the interferometric phase to vertical structure. This diversity in HoA provides complementary information for training. The available training data space is illustrated in the left panel of Fig. 2, where the observed volume coherence magnitudes $|\tilde{\gamma}_{Vol}(\kappa_z)|$ from the three acquisitions are plotted against the product $\kappa_z\, h_{\mathrm{Ref}}$, with $h_{\mathrm{Ref}}$ representing the reference forest height and $\kappa_z$ the terrain-corrected vertical wavenumber.

During training, an initial vertical reflectivity profile $f(z, a_n)$, represented by its Legendre coefficients $a_n$, is used in conjunction with $h_{\mathrm{Ref}}$ and $\kappa_z$ in (2) to predict the volume decorrelation $|\tilde{\gamma}_{\mathrm{Vol}}^{\mathrm{P}}|$. This predicted value is then compared to the observed volume coherence $|\tilde{\gamma}_{\mathrm{Vol}}(\kappa_z)|$, and the coefficients $a_n$ are iteratively updated to minimize the difference. The coefficients are estimated by a MLP that learns the mapping between the input features and the reflectivity profile.

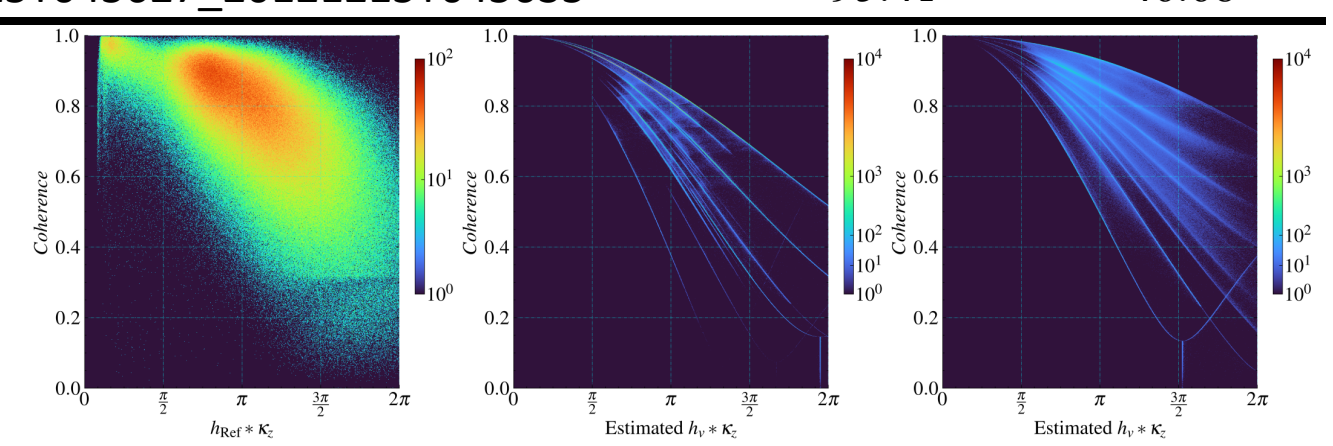

**Fig. 2.** $|\tilde{\boldsymbol{\gamma}}_{\mathbf{Vol}}(\boldsymbol{\kappa}_{\mathbf{z}})|$ vs. $\boldsymbol{\kappa}_{\mathbf{z}}\mathbf{h}_{\mathbf{v}}$ product. The plots are generated using the Lopé forest height estimates $\mathbf{h}_{\mathbf{v}}$ obtained from the inversion of all the five TanDEM-X acquisitions using two models (left model C and right model D). The colors indicate the relative number of samples and goes from dark blue (low) to dark red (high).

Once trained, the model predicts forest height in two steps. First, the input features are used to infer the vertical reflectivity profile $f(z, a_n)$. Then, this profile is employed in (2) to estimate forest height from the observed $|\tilde{\gamma}_{Vol}(\kappa_z)|$.

The resulting solution spaces from the "learned" vertical reflectivity profiles for both models are shown in the middle and right panels of Fig. 2, where $|\tilde{\gamma}_{Vol}(\kappa_z)|$ is plotted against the $\kappa_z \cdot h_v$, product where $h_v$ denotes the inverted forest height. Each profile $f(z, a_n)$ defines a unique curve in the $|\tilde{\gamma}_{Vol}(\kappa_z)|$ versus $\kappa_z \cdot h_v$, plane according to (2) associated to its solution space.

A comparison with the training data space (left panel) reveals that Model D provides a broader and more effective coverage of the data space, particularly in regions with high $\kappa_z \cdot h_v$ values and low coherence region. This indicates Model D's improved capacity to handle diverse canopy structures and a wider range of vertical wavenumbers.

**Table II:** RMSE MAE, and R² Results for Models C and D Across Different Scenes.

| Scene No | Model C | | | Model D | | |
|---|---|---|---|---|---|---|
| | RMSE [m] | MAE [m] | R² | RMSE [m] | MAE [m] | R² |
| 1 | 8.09 | 6.28 | 0.66 | 7.26 | 5.42 | 0.72 |
| 2 | 9.09 | 6.93 | 0.62 | 9.09 | 5.71 | 0.65 |
| 3 | 8.34 | 6.41 | 0.14 | 7.52 | 5.51 | 0.37 |
| 4 | 8.97 | 6.98 | 0.69 | 6.94 | 5.25 | 0.77 |
| 5 | 9.62 | 7.32 | 0.28 | 8.93 | 6.55 | 0.46 |
| **Overall** | **8.84** | **6.79** | **0.67** | **7.65** | **5.66** | **0.75** |

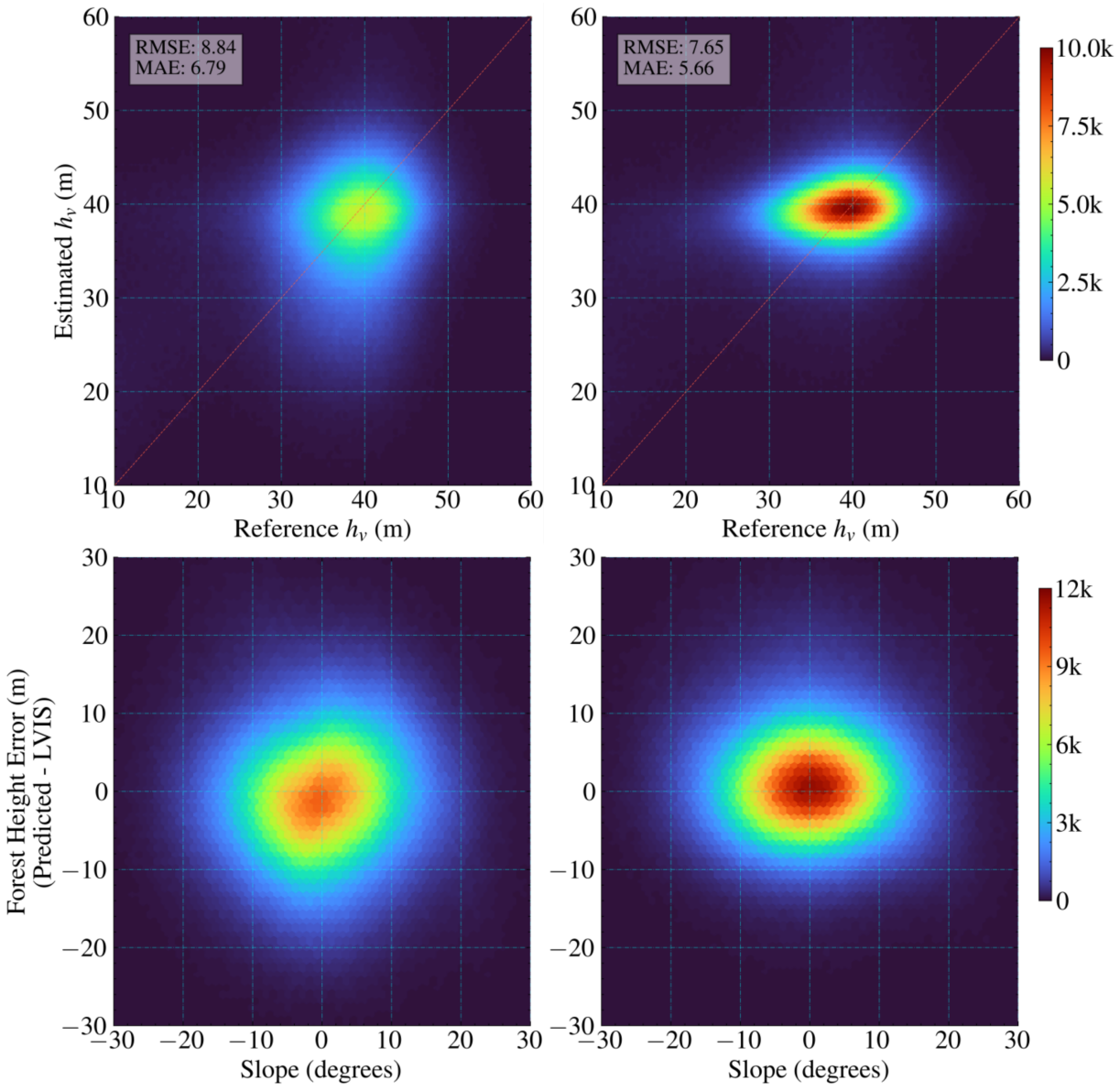

**Fig. 3.** Validation results for Models C (left) and D (right) using all five TanDEM-X acquisitions over Lopé. Top: estimated vs. reference forest height. Bottom: height residuals versus terrain slope, showing reduced bias and improved robustness for Model D.

### *A. Overall Performance*

Figure 3 (top) presents the validation plots, where the estimated forest heights $h_v$ from all five TanDEM-X acquisitions are plotted against the reference heights $\mathrm{h_{Ref}}$. Results are shown for Model C (left) and Model D (right). Performance is quantified using the mean absolute error (MAE) and root mean square error (RMSE). Model D demonstrates clear improvements over Model C. The slight underestimation observed in Model C is largely corrected in Model D, attributed to its enhanced ability to generate reflectivity profiles that better span the subspace of high $\kappa_z \cdot h_v$ and strong volume coherence $|\tilde{\gamma}_{Vol}(\kappa_z)|$.

The RMSE decreases from **8.84 m** in Model C to **7.65 m** in Model D, while the MAE improves from **6.79 m** to **5.66 m**. Table III summarizes the performance metrics across individual scenes, showing that Model D consistently outperforms Model C. The lower $R^2$ values for scenes 3 and 5 are mainly attributed to their larger HoA values despite scene 5 being included in the training set. These improvements correspond to a **13.5% reduction in RMSE** and a **16.6% reduction in MAE**. In n addition, the coefficient of determination ($R^2$) indicates that Model D ($R^2 = 0.75$) better captures the spatial variability of forest height compared to Model C ($R^2 = 0.67$). Beyond these metrics, Fig. 2 shows that Model D provides more vertical reflectivity profiles spanning a larger solution space than Model C. As a result, the hybrid framework becomes more adaptable to the forest structural variability, as also reflected by the reduced residual errors for Model D shown in the middle row of Figure 4. Consistently, false-RGB composites of the Legendre coefficients (bottom row of Figure 4) reveal that Model D captures clearer and more coherent spatial patterns than Model C, particularly at forest–savannah transitions. To further assess the model's improvement, we next analyze performance as a function of terrain slope.

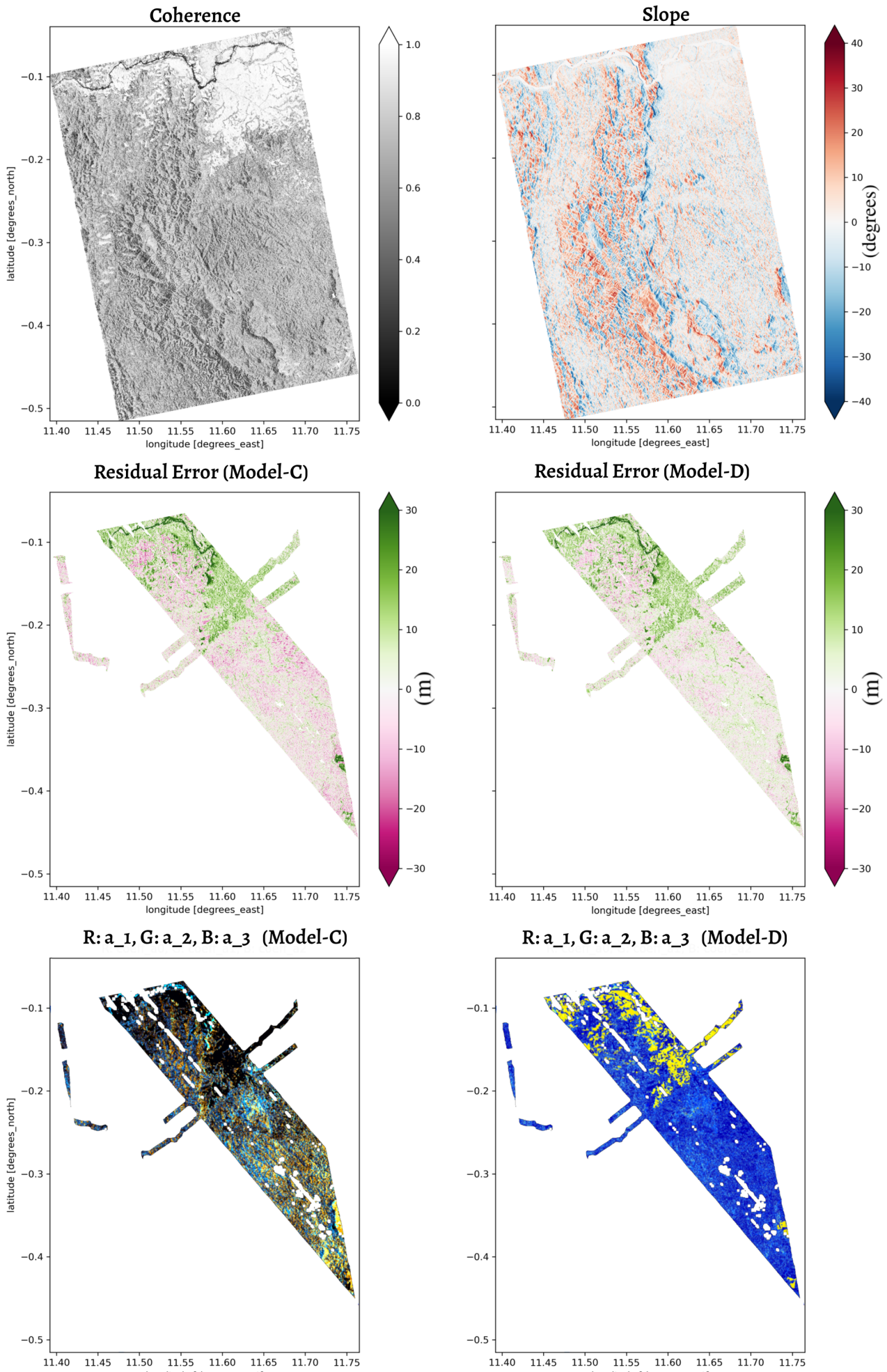

**Fig. 4.** Scene No. 2 results. Top: coherence and slope maps. Middle: residual height errors for Models C (left) and D (right). Bottom: false-RGB composites of the first three Legendre coefficients, showing clearer spatial patterns for Model D, especially across forest–savannah transitions.

### *B. Performance on Terrain Slopes*

The influence of terrain slope on model performance is analyzed next. Figure 3 (bottom) shows the residual errors (differences between estimated and reference heights) for both models across the range of terrain slopes. Model C tends to underestimate forest height at negative range slopes—those facing away from the radar—while Model D shows minimal sensitivity to slope variation.

Both models exhibit increased error variability on steeper slopes, particularly beyond ±20°, yet Model D maintains lower residuals throughout. Integrating Landsat spectral bands in Model D provides complementary information that helps resolve ambiguities arising from terrain slope and acquisition geometry. For mild slopes between −10° and +10°, Model D achieves near-zero bias and symmetrical residuals.

These results highlight the strength of the hybrid framework: the machine learning component effectively "learns" a mapping between the spectral features and the vertical reflectivity profile, compensating for underestimation in negatively sloped terrain. As a result, Model D exhibits enhanced generalization capabilities across more complex topographic conditions.

## V. Conclusion

This study proposes an extension of an established hybrid model for forest height estimation using single-baseline, single-polarization TanDEM-X interferometric coherence measurements. To enhance the model's performance with respect to height / structure and baseline / terrain slope ambiguities, the input feature space was expanded by incorporating four Landsat spectral bands. The integration of Landsat spectral information significantly improved overall performance, reducing the RMSE and MAE by 13.5% and 16.6%, respectively. It enhanced performance over complex terrain by mitigating ambiguities related to terrain slopes, acquisition geometries, and forest structural variation. The estimation bias was largely compensated for range terrain slopes ranging from $-15°$ to $+15°$.

The incorporation of multimodal data was critical in overcoming the limitations of single-source approaches. The machine learning component enabled the effective use of Landsat features by learning relationships that cannot be explicitly modeled by the physical component due to the absence of established physical relationships.

Finally, it should be noted that while this study adopts a relatively simple MLP architecture, the hybrid framework is not restricted to this choice. Other deep learning models — such as convolutional neural networks (CNNs) or transformer-based architectures with attention mechanisms—could also be considered to capture additional spatial dependencies and complex feature interactions. An MLP was chosen for its interpretability, efficiency, and suitability for multimodal integration, but future research may explore more advanced architectures.